\documentclass[letterpaper, 10 pt, conference]{ieeeconf}

\IEEEoverridecommandlockouts

\usepackage[backend=biber,
            hyperref=true,
            url=false,
            isbn=false,
            doi=false,
            backref=false,
            style=ieee,
            citestyle=numeric-comp,
            maxbibnames=99,
            minbibnames=99,
            sorting=none,
            block=none]{biblatex}
\renewcommand{\bibfont}{\small}
\usepackage[most]{tcolorbox}

\newtcolorbox{findingbox}[1]{%
  enhanced,
  colback=gray!5,
  colframe=black!70,
  arc=0mm,
  boxrule=0pt,
  leftrule=3pt,
  left=8pt, right=8pt, top=6pt, bottom=6pt,
  fonttitle=\bfseries,
  title=#1,
  coltitle=black,
  attach title to upper=\par\medskip
}

\usepackage{graphics}
\usepackage{epsfig}
\usepackage{amsmath}
\usepackage{amssymb}

\usepackage{newtxtext}
\usepackage{newtxmath}
\usepackage{bm}

\usepackage{xcolor}
\usepackage{listings}
\usepackage{booktabs}
\usepackage{tabularx}
\usepackage{array}
\usepackage{multirow}
\usepackage{colortbl}
\usepackage{makecell}
\usepackage{pbox}
\usepackage{longtable}
\usepackage{tablefootnote}
\usepackage{bigstrut}

\usepackage{graphicx}
\usepackage{subcaption}
\usepackage{wrapfig}
\usepackage{float}

\usepackage[font=small]{caption}
\usepackage[hidelinks]{hyperref}

\usepackage{xspace}
\usepackage{pifont}

\usepackage{ragged2e}
\definecolor{mahogani}{RGB}{192,64,0}
\definecolor{royalblue}{RGB}{65,105,225}

\def\ours{Sequential-EQA\xspace}

\title{\LARGE \bf
Beyond Episodic Evaluation: Memory Architectural Bottlenecks in Sequential Embodied Question Answering
}

\author{Zikui Cai$^{1,*,\dagger}$, Kaushal Janga$^{1,*}$, Tan Dat Dao$^{1,*}$, Seungjae Lee$^{1,*}$, Shivin Dass$^{2}$, Mingyo Seo$^{2}$, Kaiyu Yue$^{1}$,\\
Mintong Kang$^{3}$, Nandhu Pillai$^{1}$, Monte Hoover$^{1}$, Aadi Palnitkar$^{1}$, Ruchit Rawal$^{1}$, Ruijie Zheng$^{1}$, \\
Bo Li$^{3}$, Yuke Zhu$^{2}$, Roberto Martín-Martín$^{2}$, Tom Goldstein$^{1}$, and Furong Huang$^{1}$%
\thanks{$^{1}$University of Maryland, College Park. $^{2}$The University of Texas at Austin. $^{3}$University of Illinois Urbana-Champaign. $^{*}$Equal contribution. $^{\dagger}$Corresponding to: zikui@umd.edu. Project page: \url{https://sequential-eqa.github.io/}. Code: \url{https://github.com/jangablox/sequential-eqa}.}%
}

\begin{document}

\maketitle
\thispagestyle{empty}
\pagestyle{empty}

\begin{abstract}
Embodied question answering (EQA) is traditionally evaluated under an episodic formulation, where agents solve each task independently and reset internal state between episodes. However, real-world robots operate continuously and must accumulate, retain, and selectively reuse information acquired from prior interactions. Despite this practical requirement, the architectural mechanisms needed to support sequential memory in EQA remain underexplored. 
In this work, we investigate how different memory architectures behave when EQA agents are evaluated sequentially, with multiple questions answered in the same scene while memory is carried forward across queries. We find that simply preserving existing memory is often insufficient. Agents that retain only traversability information, such as 2D occupancy maps, remember where the robot has explored but not the visual-semantic evidence needed for later questions. Agents trained on short-horizon episodic data face a different challenge: when exposed to continuous, multi-query histories, their inherited context suffers from severe temporal mismatch, rather than forming a reusable scene representation. To overcome this architectural bottleneck, we highlight the necessity of structured, spatially grounded memory: architectures that map persistent visual observations onto metric 3D geometry preserve visual-semantic evidence in a coherent scene representation. Extensive experiments in simulated environments reveal that this form of memory breaks the accuracy-efficiency tradeoff in sequential settings, simultaneously achieving higher answer accuracy and lower navigation costs. We further validate these findings on a real-world mobile robot, demonstrating that spatially grounded visual memory is critical for enabling continuous, intelligent operation in physical environments.

\end{abstract}

\section{Introduction}

Embodied question answering (EQA)~\cite{das2018embodied} is a core AI capability that requires a physical agent to perceive, reason, and interact within an environment to answer natural language queries grounded in the physical world~\cite{anderson2018vision, thomason2020vision}. EQA serves as a foundational building block for a diverse range of real-world scenarios, ranging from domestic assistive robotics to industrial search-and-rescue and warehouse logistic, where agents must perform long-horizon tasks. Over the past several years, substantial progress has been made on EQA benchmarks and systems~\cite{wijmans2019embodied, express, majumdar2024openeqa, goatbench, yu2019multi, chi2020just, krishna2017visual, shridhar2020alfred}, enabling agents to achieve strong performance on a wide range of visually grounded questions in complex scenes. However, most existing evaluations adopt an episodic formulation, in which each question is treated as an independent task and the agent's internal state is reset between episodes~\cite{das2018embodied, wijmans2019embodied, majumdar2024openeqa, express}. This episodic assumption is increasingly misaligned with real-world deployment, where robots operate continuously and must reuse knowledge acquired from prior interactions to perform subsequent tasks efficiently.

Concretely, an EQA episode begins when an agent receives a natural-language question such as ``What color is the mug on the kitchen counter?'' while located in a partially observed 3D environment. To answer, the agent must decide where to move, collect visual observations, determine when it has gathered sufficient evidence. Traditionally, this process has relied on modular robotic pipelines for active perception, localization, mapping, and collision-free navigation~\cite{yamauchi1997frontier, thrun2002robotic, das2018embodied, wijmans2019embodied}. While recent advances in LLMs and VLMs have significantly enhanced the agent's ability to ground language and reasoning over incomplete scene information, repeating this intensive exploration loop from scratch for every subsequent question is prohibitively inefficient. As embodied agents transition toward continuous, long-horizon operation, maintaining a persistent memory becomes an unavoidable necessity to bypass redundant re-exploration. Yet, a common implicit assumption in current systems is that simply leaving an agent's memory ``on" across tasks will uniformly improve efficiency without degrading decision quality. In this work, we challenge this assumption and investigate a critical architectural bottleneck: how do different memory structures behave when transitioned from episodic to sequential EQA?

\begin{figure*}[t]
    \centering
    \begin{subfigure}[b]{0.95\linewidth}
        \centering
        \includegraphics[width=\linewidth]{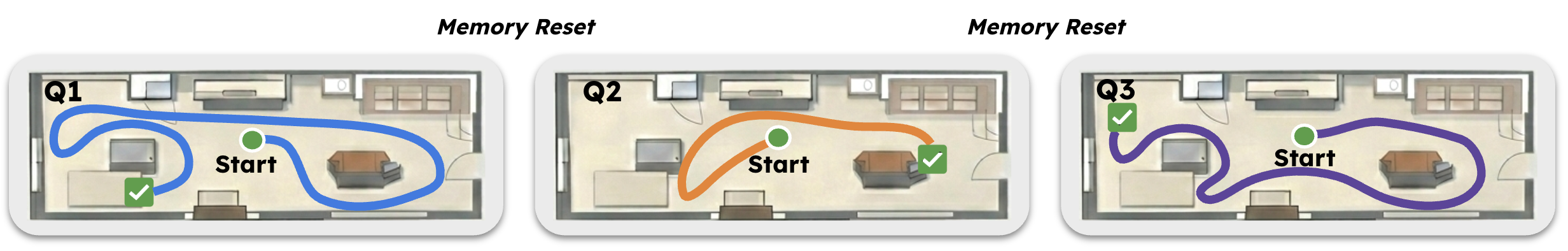}
        \caption{Episodic Evaluation}
        \label{fig:episodic}
    \end{subfigure}
    \hfill
    \begin{subfigure}[b]{0.95\linewidth}
        \centering
        \includegraphics[width=\linewidth]{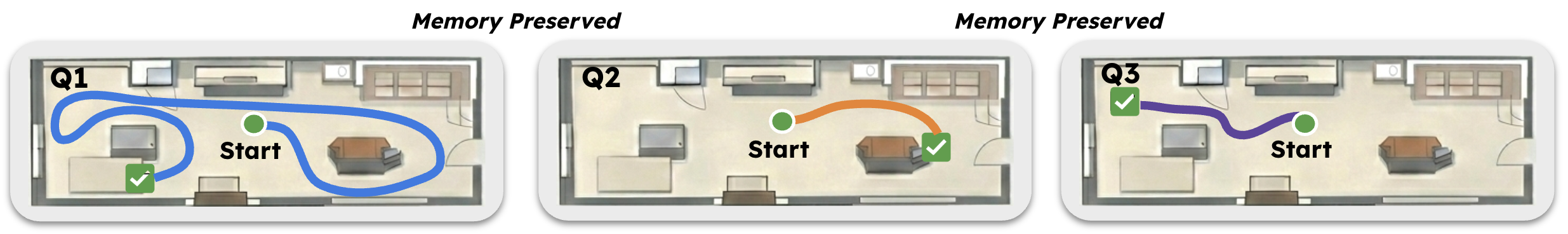}
        \caption{Sequential Evaluation}
        \label{fig:sequential}
    \end{subfigure}
    
    \caption{\textbf{Comparison between Episodic and Sequential Evaluation paradigms.} 
    (a) In \textbf{episodic evaluation}, the agent's memory is cleared after every task; for each new query (e.g., Q2), the agent must repeat the full exploration process as if the environment were novel. 
    (b) Conversely, \textbf{sequential evaluation} allows the agent to maintain a persistent memory across successive tasks. While the initial task (Q1) requires full exploration, the agent leverages its prior knowledge in subsequent tasks (Q2) to navigate directly to targets, eliminating redundant re-exploration and significantly improving efficiency.}
    \label{fig:conceptual_comparison}
\end{figure*}

To systematically study this, we introduce the Sequential-EQA evaluation protocol, illustrated in Fig.~\ref{fig:conceptual_comparison}. With minimal intervention, we convert widely used EQA benchmarks, such as OpenEQA~\cite{majumdar2024openeqa}, into a sequential evaluation setting by presenting multiple questions consecutively within the same environment and carrying the agent’s internal state across questions, while leaving the underlying environments and model parameters unchanged. We evaluate existing embodied QA methods that employ different memory mechanisms, including short-term episodic buffers and persistent spatial–semantic representations. These methods were originally designed for independent episodes; we intentionally adapt them without additional optimization to reuse memory across tasks in order to study failure modes rather than optimized performance. 

Specifically, we compare three representative memory architectures: lightweight geometric memory, implicit memory in end-to-end VLA agents, and structured 3D spatial memory. Our findings reveal that naively reusing memory exposes severe vulnerabilities in existing architectures. Agents relying on weak memory structures, such as 2D occupancy maps (e.g., ExploreEQA~\cite{ren2024explore}), fail to retain the rich semantic information required for complex sequential reasoning. Conversely, agents trained end-to-end on short-horizon episodic data (e.g., vision-language-action (VLA) approach UniNavid~\cite{zhang2024uni}) suffer from severe out-of-distribution degradation when exposed to continuous, multi-query histories. For these architectures, memory reuse substantially reduces navigation cost but causes their accuracy to either stagnate or systematically degrade due to semantic interference and overwriting.

Using Sequential-EQA, we demonstrate the necessity of structured, spatially-grounded memory. We show that 3D-Mem \cite{yang20253d}, an architecture that maps persistent visual observations to rigid 3D geometry and leverages strong Vision-Language Models (VLMs), effectively accumulates environmental knowledge over time. Unlike reactive baselines, 3D-Mem uniquely breaks the accuracy-efficiency tradeoff in sequential settings. By anchoring visual memory to spatial coordinates, it avoids catastrophic forgetting and semantic noise, simultaneously achieving higher answer accuracy and drastically lower navigation costs on subsequent queries.

To ensure our findings are not merely simulation artifacts, we reproduce our evaluation on a real mobile robot operating in physical indoor environments. The real-world deployment confirms that spatially-grounded visual memory is critical for enabling continuous, intelligent operation under realistic sensing and actuation noise.

In summary, our main contributions are:
\begin{itemize}
    \item We introduce Sequential-EQA, an evaluation protocol that transitions EQA from isolated episodes to continuous multi-query operation, measuring memory reuse efficacy without modifying underlying models or environments.
    \item We identify a critical architectural bottleneck: memory persistence does not imply knowledge accumulation, demonstrated across four architectures spanning VLM-agent and VLA paradigms.
    \item We show that spatially grounded 3D memory is necessary to break the accuracy–efficiency tradeoff, simultaneously achieving $+33.3\%$ accuracy gain and $+53.3\%$ navigation cost reduction in sequential settings.
    \item We validate on a physical quadruped robot across indoor and outdoor environments, confirming that identified bottlenecks generalize beyond simulation.
\end{itemize}

\section{Related Work}
\label{sec:related}

\noindent \textbf{Embodied Question Answering (EQA)} has emerged as a fundamental task for intelligent agents, requiring the integration of visual perception, spatial reasoning, and navigation capabilities. Early work by Das et al.~\cite{das2018embodied} introduced the first EQA benchmark using House3D environments, establishing the foundational paradigm of agents navigating to answer questions about their surroundings. This was followed by several benchmarks that expanded the scope and complexity of embodied reasoning tasks~\cite{Gordon_2018_CVPR,wijmans2019embodied,yu2019multi,majumdar2024openeqa,ren2024explore,industryeqa}.
The evolution of EQA benchmarks has primarily focused on increasing dataset scale, improving question naturalness, and expanding environmental diversity. However, by evaluating agents on isolated episodes, existing benchmarks fail to capture essential capabilities for practical deployment, where agents operate continuously, and must learn from accumulated expierence
Recent work has begun to address some aspects of continual learning in embodied AI. GOAT-Bench~\cite{goatbench} introduces lifelong navigation memory, while PartnR~\cite{partnr} incorporates task tracking across multi-agent scenarios. Long-horizon EQA has also been studied directly: Enter the Mind Palace~\cite{ginting2025mindpalace} evaluates agents that must reason and plan over extended active question-answering interactions. Our work is complementary rather than claiming that Sequential-EQA is the first long-horizon EQA setting. The distinction is that we focus on a diagnostic evaluation protocol for existing episodic EQA agents: we keep the environments, questions, and model weights fixed, then vary only whether memory persists across multiple questions in the same scene. This lets us isolate whether a memory architecture itself supports reusable knowledge accumulation, rather than measuring the performance of a method specifically designed or trained for long-horizon operation.

\noindent \textbf{Memory Architectures in Embodied AI.}
Memory plays a central role in embodied intelligence. Prior work has explored a variety of representations, for building and retaining information, including semantic and voxel maps~\cite{shafiullah2022clip}, scene graphs~\cite{chang2021comprehensive}, and other structured memory representations~\cite{gupta2017cognitive, chaplot2020object}. These representations enable localization, object search, and intra-episode reasoning by integrating observations over time. Recent systems further combine structured spatial memory with large Vision–Language Models to improve semantic grounding and exploration strategies~\cite{yokoyama2024vlfm,ren2024explore,yang20253d,zhai2025memory}.
However, most existing memory mechanisms are designed to support decision-making within a single episode. Their performance is rarely evaluated under conditions where memory must persist across multiple, temporally linked queries. Consequently, it remains unclear whether these architectures encode environmental knowledge in a form that supports long-horizon reuse, or whether they primarily function as short-term navigation aids. Our work isolates this question by transitioning episodically trained agents to sequential evaluation without modifying their training procedures.

\section{Sequential Evaluation Protocol}

Traditional episodic evaluation treats each question as independent: the agent is reset and no internal memory, maps, or perceptual representations carry over. This simplifies evaluation but ignores whether agents can accumulate knowledge from previous interactions. We instead evaluate sequences of questions in the same environment continuously without resetting internal state, exposing whether memory persistence actually improves later answers.

\subsection{Problem Formulation: Episodic vs. Sequential}

We define the EQA task within an environment $\mathcal{E}$ containing an ordered set of $N$ questions $\mathcal{Q} = \{q_1, q_2, \dots, q_N\}$. The agent is characterized by a policy $\pi$ and an internal state (memory) $m \in \mathcal{M}$, which may include spatial maps, latent embeddings, or navigation history.

\textbf{Episodic Evaluation:} In the standard episodic protocol, each question $q_i \in \mathcal{Q}$ is treated as a disjoint Markov decision process (MDP). The agent's memory is re-initialized to a null state $m_0 = \emptyset$ at the start of every question. The trajectory $\tau_i$ for question $q_i$ is generated as:
\begin{equation}
\tau_i \sim \pi(q_i, m_0).
\label{eq:episodic_trajectory}
\end{equation}
This ensures the agent ignores any environmental knowledge gained in previous trials.

\textbf{Sequential Evaluation:} In the sequential protocol, the memory is accumulated over the set of questions $\mathcal{Q}$. If $m_i$ is the memory at the terminal state for question $q_i$, then the trajectory for question $q_{i+1}$ is generated as 
\begin{equation}
\tau_{i+1} \sim \pi(q_{i+1}, m_i).
\label{eq:sequential_trajectory}
\end{equation}
This allows the policy to leverage accumulated environmental context.

\subsection{Sequentializing an Existing Benchmark}
To isolate memory reuse, we group questions from the same environment $\mathcal{E}$ into a sequence $\mathcal{S}$. Given the original dataset $\mathcal{D}_{\text{episodic}} = \{(q_j, \mathcal{E}_j)\}_{j=1}^M$, the sequential version is:
\begin{equation}
\mathcal{D}_{\text{seq}} = \{ (\mathcal{S}_k, \mathcal{E}_k) \}_{k=1}^{K}.
\label{eq:sequential_dataset}
\end{equation}
Each $\mathcal{S}_k = (q_{k,1}, \dots, q_{k,N_k})$ contains all selected questions from scene $\mathcal{E}_k$; thus $N_k$ varies by scene. Sequence order is randomly shuffled with a fixed seed and held constant for all agents and episodic/sequential comparisons. We will release the generated sequences, split code, and evaluation configuration for reproducibility.

\subsection{Evaluation Metrics}

We quantify the efficacy of memory retention through both task success and navigational efficiency. Let $y_{k,i}$ be the success indicator (binary or real-valued score) for the $i$-th question in sequence $\mathcal{S}_k$, and $L_{k,i}$ be the geodesic path length (in meters) traversed during that question.

\noindent\textbf{Accuracy metrics.} Success Rate ($SR$) is the mean accuracy under episodic resets:
\begin{equation}
SR = \frac{1}{\sum_{k=1}^{K} N_k} \sum_{k=1}^{K} \sum_{i=1}^{N_k} y_{k,i}
\quad \text{s.t.} \quad m_{0,k,i} = \emptyset.
\label{eq:success_rate}
\end{equation}
Success Rate with Memory ($SR_{\text{mem}}$) is the mean accuracy under state persistence:
\begin{equation}
SR_{\text{mem}} = \frac{1}{\sum_{k=1}^{K} N_k} \sum_{k=1}^{K} \sum_{i=1}^{N_k} y_{k,i}
\quad \text{s.t.} \quad m_{0,k,i} = m_{T_{k,i-1},k,i-1}.
\label{eq:success_rate_memory}
\end{equation}
Memory Advantage ($MA$) is the absolute performance gain of using sequential memory:
\begin{equation}
MA = SR_{\text{mem}} - SR.
\label{eq:memory_advantage}
\end{equation}

\noindent\textbf{Efficiency metrics.} Path Length ($PL$) is the total navigation cost required without information reuse:
\begin{equation}
PL = \sum_{k=1}^{K} \sum_{i=1}^{N_k} L_{k,i}
\quad \text{s.t.} \quad m_{0,k,i} = \emptyset.
\label{eq:path_length}
\end{equation}
Path Length with Memory ($PL_{\text{mem}}$) is the total cost when leveraging prior exploration:
\begin{equation}
PL_{\text{mem}} = \sum_{k=1}^{K} \sum_{i=1}^{N_k} L_{k,i}
\quad \text{s.t.} \quad m_{0,k,i} = m_{T_{k,i-1},k,i-1}.
\label{eq:path_length_memory}
\end{equation}
Step Advantage ($SA$) is the normalized efficiency gain, expressed as a percentage:
\begin{equation}
SA = \left( \frac{PL - PL_{\text{mem}}}{PL} \right) \times 100\%.
\label{eq:step_advantage}
\end{equation}

\section{Memory Representation Paradigms in EQA}
\label{sec:memory_mechanism}

Existing EQA systems differ in how they connect perception, memory, and action. \textbf{VLM-agent methods} keep a pretrained VLM frozen and collect structured observations for inference, using memories such as 2D maps~\cite{ren2024explore}, dense semantic libraries~\cite{zhai2025memory}, or 3D reconstructions~\cite{yang20253d}. This makes memory design modular, but also means that retrieval quality and representation structure strongly determine performance. \textbf{VLA methods} instead fine-tune the VLM to output actions or waypoints end-to-end, with memory implicit in short-horizon hidden states. This tight perception-action coupling can be effective within the training distribution, but limits memory to the horizon seen during training, which becomes vulnerable under sequential evaluation.

\begin{table*}[t]
\centering
\caption{Comparison of memory paradigms in embodied EQA. We categorize each method by its architectural paradigm, the specific content stored, and the primary retrieval mechanism employed.}
\label{tab:memory_paradigms}
\small
\setlength{\tabcolsep}{6pt}
\renewcommand{\arraystretch}{1.3}

\begin{tabularx}{\textwidth}{@{} l c X X @{}}
\toprule
\textbf{Method} & \textbf{Paradigm} & \textbf{What is Stored} & \textbf{Retrieval Mechanism} \\
\midrule

Explore-EQA 
& VLM-Agent 
& Occupancy + frontier scores 
& Frontier ranking by VLM \\

\midrule

MemoryEQA 
& VLM-Agent 
& RGB + pose + language description + embedding 
& Entropy-based adaptive retrieval \\

\midrule

3D-Mem 
& VLM-Agent 
& 3D point cloud + visual embeddings 
& Spatially indexed 3D lookup \\

\midrule

UniNavid 
& VLA 
& Transformer hidden state 
& Implicit attention \\

\bottomrule
\end{tabularx}
\end{table*}

\subsection{Evaluated Agents}
We select four representative agents to span these paradigm boundaries; their key structural differences in memory, retrieval, and adaptation are summarized in Table~\ref{tab:memory_paradigms}.

\noindent \textbf{VLM-agent with geometric memory.} 
ExploreEQA \cite{ren2024explore} maintains a 2D occupancy map encoding traversable space and frontier scores. A frozen VLM scores candidate frontiers for semantic relevance given the current question, guiding exploration toward likely answer locations. Because the map stores only geometric occupancy rather than object-level appearance or semantic attributes, it effectively remembers where the agent has been but not what was observed there.

\noindent \textbf{VLM-agent with dense episodic semantic memory.}
MemoryEQA~\cite{zhai2025memory} builds an explicit semantic library in which each entry binds an RGB frame to its spatial pose, a natural language description, and a feature embedding. At inference time, an entropy-based retrieval strategy selects the most relevant entries for the current query. While semantically rich, entries are stored as independent pose-tagged events rather than fused into a global spatial structure, which limits coherent reasoning across viewpoints.

\noindent \textbf{VLM-agent with structured 3D spatial-semantic memory.}
3D-Mem~\cite{yang20253d} constructs a persistent metric reconstruction of the environment, binding visual embeddings directly to 3D coordinates. Observations from different viewpoints are fused into a spatially consistent representation, preserving both layout and object identity. Retrieval operates over structured 3D geometry, enabling the agent to reason about spatial relationships explicitly and accumulate knowledge compositionally across queries.

\noindent \textbf{VLA with implicit latent memory.}
UniNavid~\cite{zhang2024uni} fine-tunes a transformer-based VLM to directly predict navigation waypoints from a short window of sequential RGB frames and a language query. Memory is implicit in the model's attention state across this window. Because training always begins from a fresh episode, the model has no mechanism to consolidate observations across query boundaries, making it inherently episodic by design.

\subsection{Memory Adaptation Protocol}
To isolate the contribution of memory architecture from the confounding effects of sequential training, we adopt a minimal persistence protocol that transitions each agent from episodic to sequential evaluation without any modification to model weights or training procedures. This design choice is deliberate: our goal is not to find the best-performing sequential agent after task-specific optimization, but rather to diagnose which architectural properties intrinsically support knowledge accumulation across queries.
The protocol applies uniformly to all agents. At the boundary between consecutive questions $q_i$ and $q_{i+1}$ within a scene, we perform three operations: (1) \textbf{state inheritance}: The agent's terminal memory state, $m_{T_i,i}$, is carried forward as the initial state for the next query, $m_{0,i+1}$, with no truncation or summarization. (2) \textbf{query update}: The task-specific input is replaced with $q_{i+1}$, and any goal-directed planners are re-initialized accordingly. (3) \textbf{frozen weights}: All network parameters $\theta$ remain strictly unchanged throughout the sequence, and no uncertainty recalibration or auxiliary adaptation is applied.
This minimal intervention ensures that any performance difference between episodic and sequential conditions is attributable solely to the structure of the inherited representation, not to any learned sequential behavior. Agents that benefit from this protocol do so because their memory architecture is compositional by design; agents that fail do so because their representations were never structured to support cross-query reuse.
\section{Experimental Results and Analysis}
\label{sec:results}

\subsection{Experimental Setup}
\begin{figure}[t]
    \centering
    \includegraphics[width=0.95\linewidth]{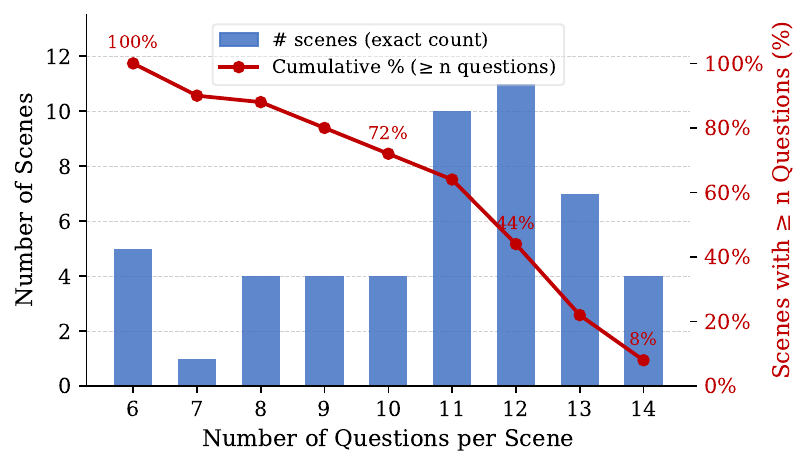}
    \caption{Distribution of questions per scene in the sequentialized OpenEQA benchmark. Bars show the number of scenes containing exactly $N_k$ questions; the overlaid line shows the cumulative percentage of scenes with at least $N_k$ questions. We evaluate all generated scene-level sequences; aggregate metrics pool all questions, while query-index analyses average over scenes with $N_k \geq i$.}
    \label{fig:dataset-stats}
\end{figure}
\noindent \textbf{Dataset.} We construct our sequential benchmark by re-grouping questions from the OpenEQA~\cite{majumdar2024openeqa} dataset by scene and evaluate the aforementioned 4 different types of representative agents. OpenEQA provides real indoor environments paired with free-form questions that require spatial and semantic reasoning. Under our sequential protocol, all selected questions belonging to the same scene are concatenated into one ordered sequence, and the agent's internal state is preserved across them without any resets. We evaluate all resulting scene-level sequences rather than truncating every scene to a fixed horizon: aggregate results pool all questions from all sequences, while per-query-index plots include, at index $i$, only scenes whose sequence length satisfies $N_k \geq i$. Figure~\ref{fig:dataset-stats} shows the distribution of sequence lengths across scenes in the resulting benchmark. The majority of scenes contain between 10 and 13 questions, with over 60\% of scenes providing sequences of at least 10 queries. The exact shuffled sequence files, split-generation code, and evaluation hyperparameters will be released with the benchmark to make the random ordering reproducible and to support future controlled comparisons.

\noindent \textbf{Foundation model and compute resources.} We use Qwen3 VL 8B-Instruct~\cite{bai2025qwen3} VLM with FB8 quantization, all experiments are run on A5000 GPUs.

\subsection{Sequential Memory Does Not Uniformly Improve Performance}
\label{sec:exp-overall-performance}

\begin{table}[t]
\centering
\caption{Performance comparison on \ours benchmark. Results show success rates under episodic (SR \%) and sequential (SR$_{\text{mem}}$ \%) conditions, memory advantage (MA \%), path lengths (PL), and step advantage (SA \%). $\uparrow$ indicates higher is better, $\downarrow$ indicates lower is better. Standard errors shown in Fig. \ref{fig:acc_efficiency_tradeoff}.}
\label{tab:overall_performance}
\small
\setlength{\tabcolsep}{3.5pt}
\begin{tabular}{lcccccc}
\toprule
\textbf{Method} & \textbf{SR $\uparrow$} & \textbf{SR$_{\text{mem}}$ $\uparrow$} & \textbf{MA $\uparrow$} & \textbf{PL $\downarrow$} & \textbf{PL$_{\text{mem}}$ $\downarrow$} & \textbf{SA $\uparrow$} \\
\midrule
ExploreEQA & 43.8 & 46.5 & 2.7 & 84.3 & 84.3 & 0.0 \\
MemoryEQA  & 61.0 & 62.4 & 1.4 & 43.6 & 43.9 & -0.5 \\
3D-Mem     & 25.5 & 58.8 & 33.3 & 5.6 & 2.6 & 53.3 \\
UniNavid   & 36.4 & 37.3 & 0.9 & 12.2 & 12.4 & -1.5 \\
\bottomrule
\end{tabular}
\end{table}

A natural expectation is that preserving internal state should reduce redundant
exploration while accumulating useful environmental knowledge. However,
Table~\ref{tab:overall_performance} shows that this assumption only partly
holds. Sequential evaluation separates two effects that are coupled in standard
episodic testing: whether memory helps the agent answer better, measured by MA,
and whether it helps the agent move less, measured by SA. The results lead to
four main takeaways.

\noindent \textbf{Accuracy and efficiency.}
Table~\ref{tab:overall_performance} shows that memory persistence alone does
not reliably improve answer quality. ExploreEQA, UniNavid, and MemoryEQA all
obtain near-zero memory advantage (MA $<3\%$), but for different architectural
reasons. ExploreEQA retains traversed geometry but not object-level semantic
evidence, so it remembers where the agent has been rather than what was
observed. UniNavid inherits recurrent context across query boundaries, although
it was trained on short episodic horizons; this makes sequential context an
out-of-distribution input rather than a reusable scene model. MemoryEQA stores
richer RGB observations, poses, descriptions, and embeddings, but these entries
accumulate as independent events rather than a globally aligned scene
representation, increasing retrieval noise as sequences grow.

In contrast, 3D-Mem improves both answer accuracy and navigation efficiency,
with $+33.3\%$ MA and $+53.3\%$ SA. This improvement is not merely trajectory
compression: the agent answers more questions correctly while navigating less.
By anchoring observations to metric 3D geometry, each query enriches a coherent
scene representation rather than adding unstructured history.

\begin{figure*}[t]
\centering
\includegraphics[width=0.95\textwidth]{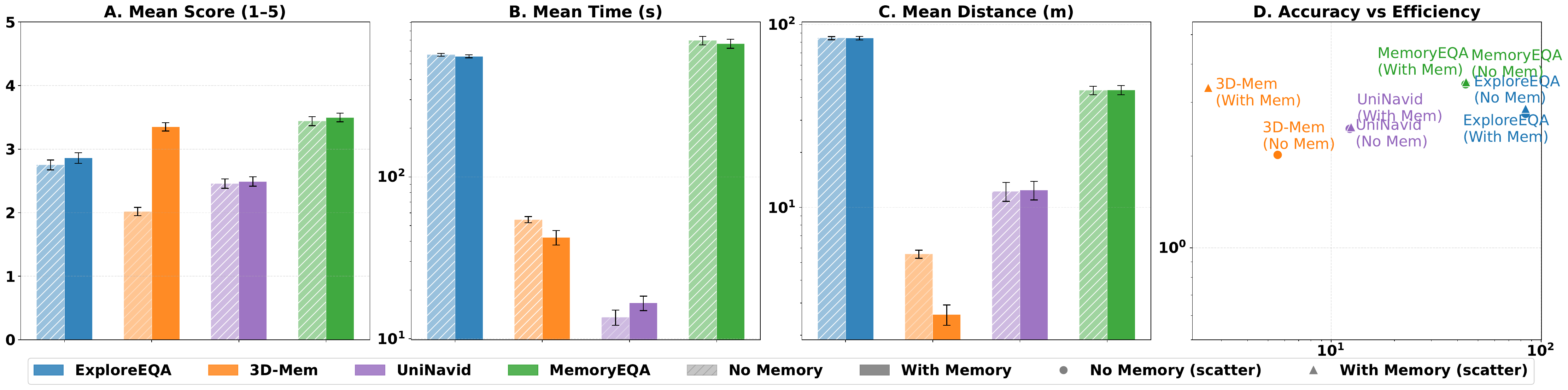}
\caption{\textbf{Accuracy--efficiency tradeoff under sequential memory reuse.} From left to right: mean answer score (1--5; $\uparrow$), mean navigation time (seconds; $\downarrow$), mean navigation distance (meters; $\downarrow$), and a joint efficiency-vs-accuracy scatter plot. Light bars denote episodic (no memory) evaluation; dark bars denote sequential (with memory) evaluation; error bars indicate standard error.}
\label{fig:acc_efficiency_tradeoff}
\end{figure*}

Figure~\ref{fig:acc_efficiency_tradeoff} visualizes the same pattern. Across
most methods, memory persistence changes navigation cost more readily than
answer quality. Only 3D-Mem moves toward the desirable region of higher
accuracy and lower navigation cost, indicating that spatially grounded memory
turns prior observations into reusable evidence rather than simply shortening
paths.

\begin{findingbox}{Key Takeaways from Sequential Evaluation}
    \small
    \textbf{Persistence $\neq$ Accumulation.} Merely retaining state across sequential queries does not guarantee successful environmental knowledge accumulation.
    
    \medskip
    \textbf{Spatial Grounding is Critical.} Only spatially grounded memory consistently translates accumulated observations into simultaneous gains in answer accuracy and navigation efficiency.
    
    \medskip
    \textbf{Decoupling of Efficiency and Success via Premature Termination.} Shorter navigation paths do not imply better reasoning; unstructured memory can bias agents to stop early in familiar regions without finding correct evidence.
    
    \medskip
    \textbf{Structured Memory Supports Cross-query Accumulation.} Structuring observations geometrically allows the agent's representation to become exponentially more useful over later queries.
\end{findingbox}

\noindent \textbf{Dynamics across query position.}
Aggregate metrics show average gains, but they do not reveal whether memory
becomes more useful as a sequence progresses. We therefore analyze accuracy and
navigation time by query position $i$.

\begin{figure*}[t]
\centering
\includegraphics[width=0.95\textwidth]{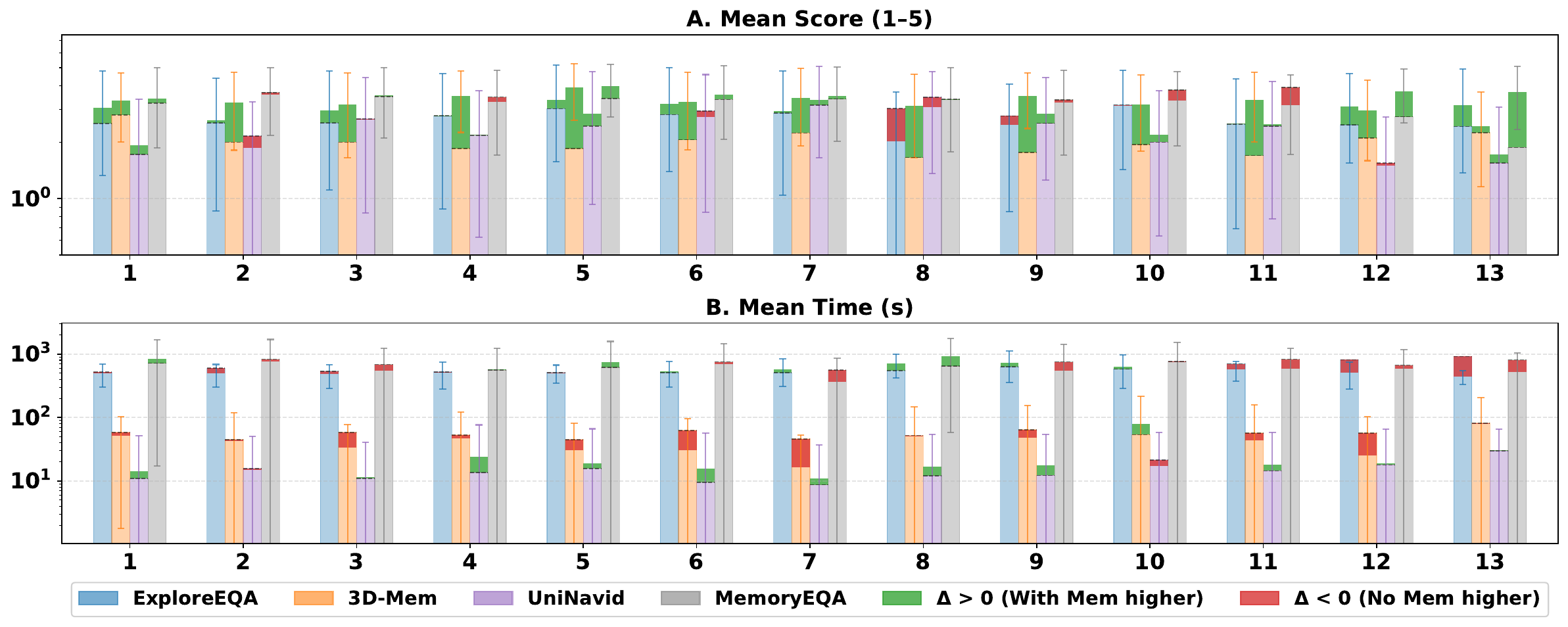}
\caption{\textbf{Per-query-index accuracy and navigation time under sequential memory reuse.} Panel A shows mean answer score (1--5; $\uparrow$, log scale); Panel B shows mean navigation time (seconds; $\downarrow$, log scale). The x-axis is query position $i$ within a scene sequence; each position averages over all sequences with $N_k \geq i$, not only sequences of length 13. $\Delta$ denotes sequential minus episodic performance and is color-coded by sign: green for $\Delta>0$ and red for $\Delta<0$. Because metric direction differs across panels, green indicates improvement only for Panel A, whereas red/negative $\Delta$ in Panel B indicates an improvement because lower time is better.}
\label{fig:score_vs_qidx}
\end{figure*}

Figure~\ref{fig:score_vs_qidx} averages each position over scenes with
$N_k \geq i$, so later points use the subset of longer sequences. Navigation
time generally decreases under memory reuse, since even weak memory can reduce
repeated exploration or bias the agent toward familiar regions. However, lower
navigation cost is not sufficient evidence of knowledge accumulation: it may
also reflect premature stopping or over-reliance on incomplete history.

Accuracy reveals this distinction. ExploreEQA, UniNavid, and MemoryEQA show no
stable improvement across query positions; gains and losses alternate as history
accumulates. This suggests that their retained state does not monotonically
improve scene understanding. Only 3D-Mem shows sustained positive accuracy
$\Delta$ across query positions, consistent with a representation that becomes
more complete over time. Taken together, the aggregate and per-query analyses
show that useful sequential memory must be both persistent and spatially
compositional.

\subsection{Qualitative Analysis}

\begin{figure}[t]
    \centering
    \begin{subfigure}{\linewidth}
        \centering
        \includegraphics[width=\linewidth]{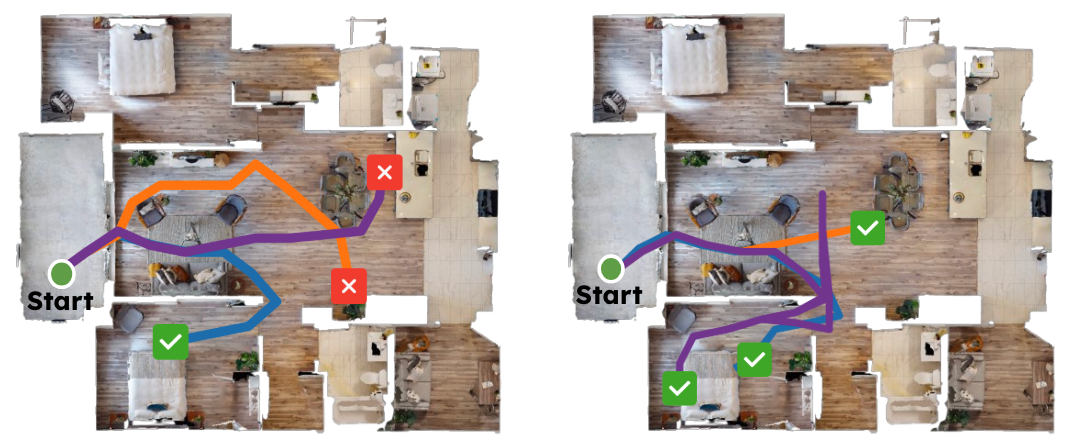}
        \caption{\textbf{3D-Mem} trajectories on a representative MP3D scene. Under episodic evaluation, the agent re-explores from scratch for each query, answering only 1/3 correctly. Persistent 3D memory enables more direct navigation and correct answers for all 3 questions.}
    \end{subfigure}
    
    \vspace{0.1in}
    
    \begin{subfigure}{\linewidth}
        \centering
        \includegraphics[width=\linewidth]{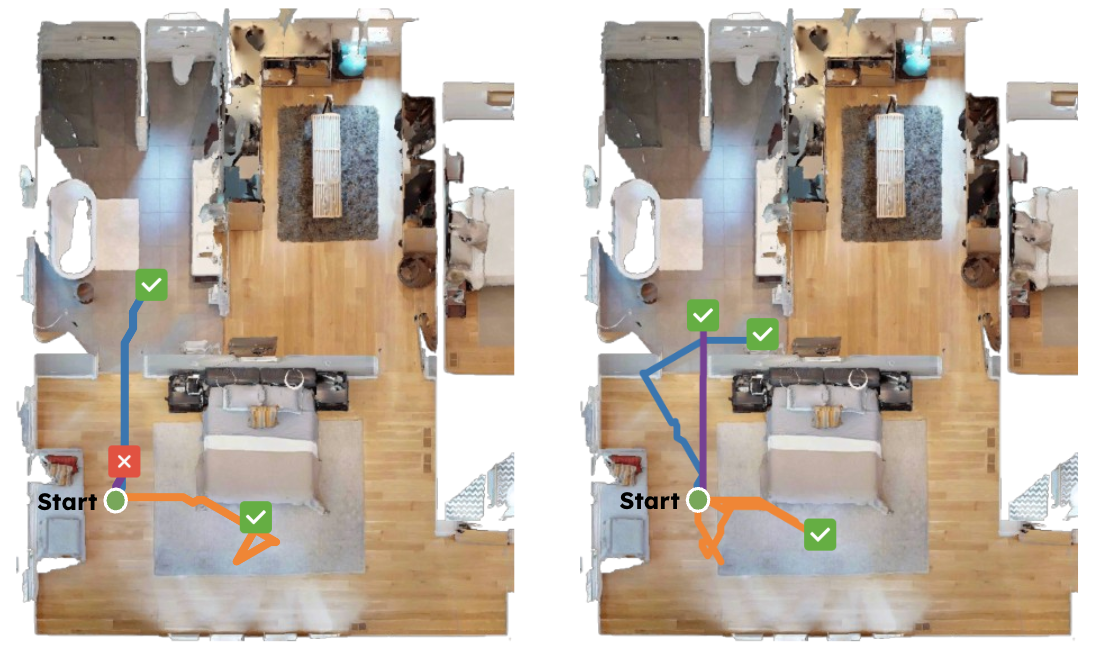}
        \caption{\textbf{UniNaviD} trajectories on another MP3D scene. The agent answers 2/3 questions correctly episodically; sequential state inheritance corrects Q2, yielding 3/3. Unlike 3D-Mem, navigation efficiency changes minimally between settings.}
    \end{subfigure}
    
    \caption{Qualitative comparison of agent trajectories under episodic (left) and sequential (right) evaluation for three consecutive questions (Q1: blue, Q2: orange, Q3: purple).}
    \label{fig:path_comparison}
\end{figure}
    
Figure~\ref{fig:path_comparison} provides a concrete illustration of the behavioral difference between episodic and sequential evaluation on a representative scene from the benchmark, using 3D-Mem as the focal architecture. The left panel shows three episodic trajectories — one per query, each beginning from the same start position with no inherited state. The agent re-explores large portions of the apartment for every question, accumulating substantial redundant path length across the sequence. Despite this exploration, it answers only 1 of 3 questions correctly: when asked "What color is the smoke detector?", it returns an answer about a coat closet — a symptom of incomplete observation coverage within the allocated episode budget.

The right panel shows the same three queries under sequential evaluation. With a persistent 3D map accumulated from prior queries, the agent navigates directly to previously observed regions rather than re-exploring from scratch. Path length drops sharply across Q2 and Q3, consistent with the aggregate SA of $53.3\%$ reported in Table~\ref{tab:overall_performance}. Critically, accuracy improves alongside efficiency: all three questions are answered correctly, including the smoke detector query, which the agent now answers correctly by drawing on observations already fused into its 3D representation from the first query's exploration.

This example illustrates the core mechanism behind 3D-Mem's sequential advantage. The improvement is not simply that the agent travels less — it is that prior exploration leaves a spatially consistent record that later queries can retrieve from directly. Efficiency and accuracy improve together because they share the same cause: a representation that accumulates observations compositionally rather than discarding them at episode boundaries.
\section{Real-Robot Validation}
\label{sec:real_world}

To verify that our simulated findings reflect physical deployment conditions, we evaluate all four methods on a real mobile robot in diverse environments. These experiments test whether the simulated bottlenecks --- distributional shift under sequential context and lack of structured spatial memory --- persist under sensing noise, actuation error, and environmental variability.

\subsection{Experimental Setup}

We deploy a Unitree Go2 quadruped with an Intel RealSense D435i depth camera and onboard LiDAR L2. ExploreEQA, MemoryEQA, 3D-Mem and UniNavid are all evaluated on both an episodic and a sequential trial per environment. All trials consist of five questions, spanning object identification, counting, and localization. Additionally, every trial resets the robot to the same start location before each query; however, the sequential trials preserved state across the five spatially dependent queries. The environments tested are a furnished indoor lab space, an open lobby, and a long hallway. UniNavid was also tested on an outdoor rooftop terrace.

\subsection{Results and Analysis}

\begin{figure}[t]
    \centering
    \begin{subfigure}{\linewidth}
        \centering
        \includegraphics[width=\linewidth]{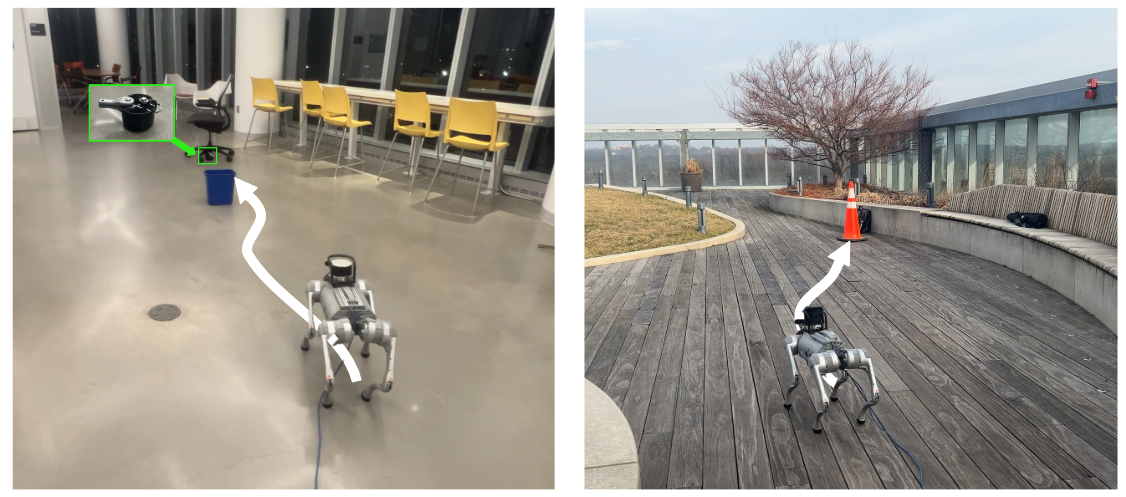}
        \caption{Question 1: Indoor ("What color is the pot next to the black chair?") and Outdoor ("What's the color of the cone?")}
    \end{subfigure}
    
    \vspace{0.1in}
    
    \begin{subfigure}{\linewidth}
        \centering
        \includegraphics[width=\linewidth]{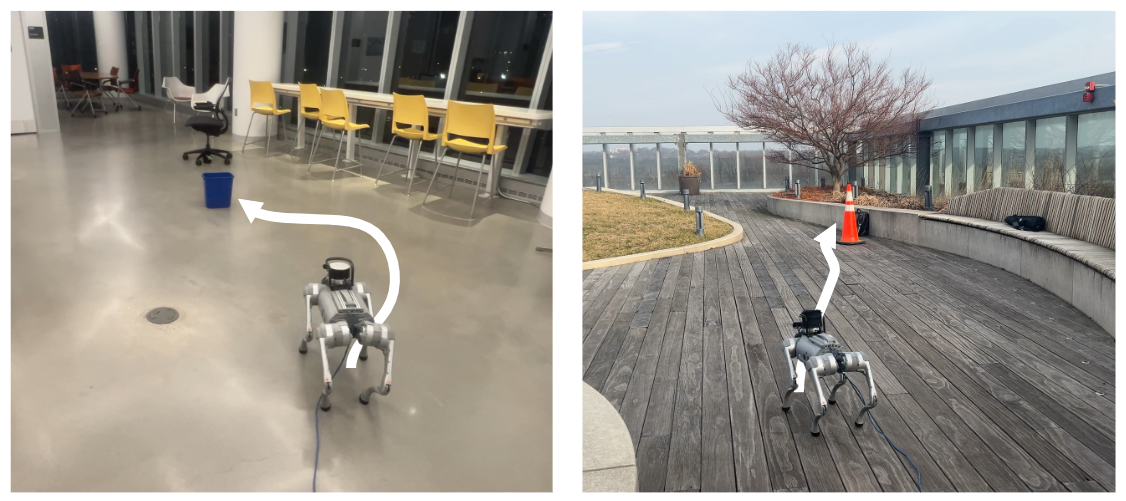}
        \caption{Question 2: Indoor ("How many yellow chairs are in this room?") and Outdoor ("What's the object behind the cone?")}
    \end{subfigure}

    \caption{\textbf{Real-world deployment of the Unitree Go2.} In both environments, UniNavid fails to leverage previous knowledge. In the indoor setting (left), the agent re-navigates to the chair for the second question as if the first encounter never occurred. Similarly, in the outdoor setting (right), the agent repeats its navigation pattern for the second query. In both cases, the agent exhibits erratic rotation and fails to answer correctly, confirming that the \textbf{sequential memory reuse bottleneck} generalizes to physical deployment.}
    \label{fig:real_world_failure}
\end{figure}

Figure~\ref{fig:real_world_failure} shows the UniNavid in both indoor and outdoor environments. UniNavid achieved $15\%$ accuracy in both episodic and sequential settings across all scenes. Failure severity is similar in both indoor and outdoor settings. 3D-Mem stayed consistent with its performance in simulation, improving its success rate from $20\%$ to $40\%$ when switching from episodic to sequential. ExploreEQA actually had a decrease in accuracy from $33\%$ to $26\%$; however, that could be due to the small test size. Finally, MemoryEQA improved from $40\%$ to $47\%$ accuracy when maintaining memory.

These results indicate that the simulation trend is not an artifact of idealized dynamics. Physical sensing noise and actuation error amplify weak memory architectures: noisy observations reduce evidence quality, and drift accumulates over longer trajectories. Without a structured map, the agent has no stable reference for deciding whether a later query can be answered from prior evidence. Structured spatial memory is therefore critical for deployable continuous EQA agents.

\section{Conclusion}
\label{sec:conclusion}

We introduced Sequential-EQA, an evaluation protocol that transitions embodied question answering from isolated episodes to continuous, multi-query evaluation. Our analysis reveals a critical architectural gap: memory persistence does not guarantee knowledge accumulation. While most paradigms fail to translate retained state into better performance due to semantic interference or a lack of spatial grounding, structured 3D spatial memory successfully breaks this bottleneck. By anchoring visual-semantic evidence to metric geometry, it eliminates redundant exploration and prevents forgetting. Our findings suggest that advancing sequential agents requires building better spatial memory structures rather than simply expanding memory capacity. By exposing these hidden failure modes in memory reuse, Sequential-EQA provides the community with a framework to bridge the gap between simulation and physical deployment. Ultimately, resolving this memory-efficiency bottleneck is the key to unlocking truly continuous, lifelong embodied assistants capable of operating reliably in human environments over days and weeks, rather than isolated, minutes-long episodes. We hope this work motivates the next generation of memory architectures that can operate continuously and reliably in the physical world.

\renewcommand{\baselinestretch}{1.03} 
\renewcommand*{\bibfont}{\footnotesize}
\printbibliography

@inproceedings{shafiullah2022clip,
  title={{CLIP-Fields}: Weakly Supervised Semantic Fields for Robotic Memory},
  author={Shafiullah, Nur Muhammad Mahi and Paxton, Chris and Pinto, Lerrel and Chintala, Soumith and Szlam, Arthur},
  booktitle={Proceedings of Robotics: Science and Systems},
  address={Daegu, Republic of Korea},
  month={7},
  doi={10.15607/RSS.2023.XIX.074},
  year={2023}
}

@article{chang2021comprehensive,
  title={A comprehensive survey of scene graphs: Generation and application},
  author={Chang, Xiaojun and Ren, Pengzhen and Xu, Pengfei and Li, Zhihui and Chen, Xiaojiang and Hauptmann, Alex},
  journal={IEEE Transactions on Pattern Analysis and Machine Intelligence},
  volume={45},
  number={1},
  pages={1--26},
  year={2021},
}

@article{chaplot2020object,
  title={Object goal navigation using goal-oriented semantic exploration},
  author={Chaplot, Devendra Singh and Gandhi, Dhiraj Prakashchand and Gupta, Abhinav and Salakhutdinov, Russ R},
  journal={Advances in Neural Information Processing Systems},
  volume={33},
  pages={4247--4258},
  year={2020}
}

@inproceedings{chi2020just,
  title={Just ask: An interactive learning framework for vision and language navigation},
  author={Chi, Ta-Chung and Shen, Minmin and Eric, Mihail and Kim, Seokhwan and Hakkani-Tur, Dilek},
  booktitle={AAAI Conference on Artificial Intelligence},
  volume={34},
  number={03},
  pages={2459--2466},
  year={2020}
}

@article{krishna2017visual,
  title={Visual Genome: Connecting Language and Vision Using Crowdsourced Dense Image Annotations},
  author={Krishna, Ranjay and Zhu, Yuke and Groth, Oliver and Johnson, Justin and Hata, Kenji and Kravitz, Joshua and Chen, Stephanie and Kalantidis, Yannis and Li, Li-Jia and Shamma, David A. and Bernstein, Michael S. and Fei-Fei, Li},
  journal={International Journal of Computer Vision},
  volume={123},
  number={1},
  pages={32--73},
  year={2017}
}

@inproceedings{shridhar2020alfred,
  title={Alfred: A benchmark for interpreting grounded instructions for everyday tasks},
  author={Shridhar, Mohit and Thomason, Jesse and Gordon, Daniel and Bisk, Yonatan and Han, Winson and Mottaghi, Roozbeh and Zettlemoyer, Luke and Fox, Dieter},
  booktitle={IEEE/CVF Conference on Computer Vision and Pattern Recognition},
  pages={10740--10749},
  year={2020}
}

@inproceedings{anderson2018vision,
  title={Vision-and-language navigation: Interpreting visually-grounded navigation instructions in real environments},
  author={Anderson, Peter and Wu, Qi and Teney, Damien and Bruce, Jake and Johnson, Mark and S{\"u}nderhauf, Niko and Reid, Ian and Gould, Stephen and Van Den Hengel, Anton},
  booktitle={IEEE/CVF Conference on Computer Vision and Pattern Recognition},
  pages={3674--3683},
  year={2018}
}

@inproceedings{thomason2020vision,
  title={Vision-and-dialog navigation},
  author={Thomason, Jesse and Murray, Michael and Cakmak, Maya and Zettlemoyer, Luke},
  booktitle={Conference on Robot Learning},
  pages={394--406},
  year={2020},
}

@inproceedings{yang20253d,
  title={3D-mem: 3D scene memory for embodied exploration and reasoning},
  author={Yang, Yuncong and Yang, Han and Zhou, Jiachen and Chen, Peihao and Zhang, Hongxin and Du, Yilun and Gan, Chuang},
  booktitle={IEEE/CVF Conference on Computer Vision and Pattern Recognition},
  pages={17294--17303},
  year={2025}
}

@article{zhai2025memory,
  title={Memory-Centric Embodied Question Answer},
  author={Zhai, Mingliang and Gao, Zhi and Wu, Yuwei and Jia, Yunde},
  journal={arXiv preprint arXiv:2505.13948},
  year={2025}
}

@inproceedings{ren2024explore,
  title={Explore until confident: Efficient exploration for embodied question answering},
  author={Ren, Allen Z and Clark, Jaden and Dixit, Anushri and Itkina, Masha and Majumdar, Anirudha and Sadigh, Dorsa},
  booktitle={Proceedings of Robotics: Science and Systems},
  address={Delft, Netherlands},
  month={7},
  doi={10.15607/RSS.2024.XX.089},
  year={2024}
}

@inproceedings{ginting2025mindpalace,
  title={Enter the Mind Palace: Reasoning and Planning for Long-term Active Embodied Question Answering},
  author={Ginting, Muhammad Fadhil and Kim, Dong-Ki and Meng, Xiangyun and Reinke, Andrzej Marek and Krishna, Bandi Jai and Kayhani, Navid and Peltzer, Oriana and Fan, David and Shaban, Amirreza and Kim, Sung-Kyun and Kochenderfer, Mykel and Agha-mohammadi, Ali-akbar and Omidshafiei, Shayegan},
  booktitle={Proceedings of the 9th Conference on Robot Learning},
  series={Proceedings of Machine Learning Research},
  volume={305},
  pages={5072--5106},
  publisher={PMLR},
  year={2025}
}

@inproceedings{yokoyama2024vlfm,
  title={Vlfm: Vision-language frontier maps for zero-shot semantic navigation},
  author={Yokoyama, Naoki and Ha, Sehoon and Batra, Dhruv and Wang, Jiuguang and Bucher, Bernadette},
  booktitle={IEEE International Conference on Robotics and Automation},
  pages={42--48},
  year={2024},
}

@inproceedings{das2018embodied,
  title={Embodied question answering},
  author={Das, Abhishek and Datta, Samyak and Gkioxari, Georgia and Lee, Stefan and Parikh, Devi and Batra, Dhruv},
  booktitle={IEEE/CVF Conference on Computer Vision and Pattern Recognition},
  year={2018}
}

@inproceedings{partnr,
  title={PARTNR: A Benchmark for Planning and Reasoning in Embodied Multi-Agent Tasks},
  author={Chang, Matthew and Chhablani, Gunjan and Clegg, Alexander and Cote, Mikael Dallaire and Desai, Ruta and Hlavac, Michal and Karashchuk, Vladimir and Krantz, Jacob and Mottaghi, Roozbeh and Parashar, Priyam and Patki, Siddharth and Prasad, Ishita and Puig, Xavier and Rai, Akshara and Ramrakhya, Ram and Tran, Daniel and Truong, Joanne and Turner, John M. and Undersander, Eric and Yang, Tsung-Yen},
  booktitle={International Conference on Learning Representations},
  year={2025}
}

@inproceedings{gupta2017cognitive,
  title={Cognitive mapping and planning for visual navigation},
  author={Gupta, Saurabh and Davidson, James and Levine, Sergey and Sukthankar, Rahul and Malik, Jitendra},
  booktitle={IEEE/CVF Conference on Computer Vision and Pattern Recognition},
  pages={2616--2625},
  year={2017}
}

@inproceedings{yu2019multi,
  title={Multi-target embodied question answering},
  author={Yu, Licheng and Chen, Xinlei and Gkioxari, Georgia and Bansal, Mohit and Berg, Tamara L and Batra, Dhruv},
  booktitle={IEEE/CVF Conference on Computer Vision and Pattern Recognition},
  pages={6309--6318},
  year={2019}
}

@inproceedings{express,
title={Beyond the Destination: A Novel Benchmark for Exploration-Aware Embodied Question Answering},
author={Jiang, Kaixuan and Liu, Yang and Chen, Weixing and Luo, Jingzhou and Chen, Ziliang and Pan, Ling and Li, Guanbin and Lin, Liang},
year={2025},
booktitle={IEEE/CVF International Conference on Computer Vision}
}

@inproceedings{industryeqa,
  title={IndustryEQA: Pushing the Frontiers of Embodied Question Answering in Industrial Scenarios},
  author={Li, Yifan and Chen, Yuhang and Dao, Anh and Li, Lichi and Cai, Zhongyi and Tan, Zhen and Chen, Tianlong and Kong, Yu},
  booktitle={Advances in Neural Information Processing Systems},
  year={2025}
}

@InProceedings{Gordon_2018_CVPR,
author = {Gordon, Daniel and Kembhavi, Aniruddha and Rastegari, Mohammad and Redmon, Joseph and Fox, Dieter and Farhadi, Ali},
title = {IQA: Visual Question Answering in Interactive Environments},
booktitle = {IEEE/CVF Conference on Computer Vision and Pattern Recognition},
year = {2018}
}

@InProceedings{wijmans2019embodied,
author = {Wijmans, Erik and Datta, Samyak and Maksymets, Oleksandr and Das, Abhishek and Gkioxari, Georgia and Lee, Stefan and Essa, Irfan and Parikh, Devi and Batra, Dhruv},
title = {Embodied Question Answering in Photorealistic Environments With Point Cloud Perception},
booktitle = {IEEE/CVF Conference on Computer Vision and Pattern Recognition},
year = {2019}
}

@InProceedings{majumdar2024openeqa,
author = {Majumdar, Arjun and Ajay, Anurag and Zhang, Xiaohan and Putta, Pranav and Yenamandra, Sriram and Henaff, Mikael and Silwal, Sneha and Mcvay, Paul and Maksymets, Oleksandr and Arnaud, Sergio and Yadav, Karmesh and Li, Qiyang and Newman, Ben and Sharma, Mohit and Berges, Vincent and Zhang, Shiqi and Agrawal, Pulkit and Bisk, Yonatan and Batra, Dhruv and Kalakrishnan, Mrinal and Meier, Franziska and Paxton, Chris and Sax, Alexander and Rajeswaran, Aravind},
title = {OpenEQA: Embodied Question Answering in the Era of Foundation Models},
booktitle = {IEEE/CVF Conference on Computer Vision and Pattern Recognition},
year = {2024}
}

@InProceedings{goatbench, author = {Khanna, Mukul and Ramrakhya, Ram and Chhablani, Gunjan and Yenamandra, Sriram and Gervet, Theophile and Chang, Matthew and Kira, Zsolt and Chaplot, Devendra Singh and Batra, Dhruv and Mottaghi, Roozbeh}, title = {GOAT-Bench: A Benchmark for Multi-Modal Lifelong Navigation}, booktitle = {IEEE/CVF Conference on Computer Vision and Pattern Recognition}, year = {2024},
pages = {16373--16383}
}

@inproceedings{zhang2024uni,
  title={{Uni-NaVid}: A Video-based Vision-Language-Action Model for Unifying Embodied Navigation Tasks},
  author={Zhang, Jiazhao and Wang, Kunyu and Wang, Shaoan and Li, Minghan and Liu, Haoran and Wei, Songlin and Wang, Zhongyuan and Zhang, Zhizheng and Wang, He},
  booktitle={Proceedings of Robotics: Science and Systems},
  address={Los Angeles, CA, USA},
  month={6},
  doi={10.15607/RSS.2025.XXI.013},
  year={2025}
}

@article{bai2025qwen3,
  title={{Qwen3-VL} Technical Report},
  author={{Qwen Team}},
  journal={arXiv preprint arXiv:2511.21631},
  year={2025}
}

@inproceedings{yamauchi1997frontier,
  title={A frontier-based approach for autonomous exploration},
  author={Yamauchi, Brian},
  booktitle={Proceedings 1997 IEEE International Symposium on Computational Intelligence in Robotics and Automation CIRA'97.'Towards New Computational Principles for Robotics and Automation'},
  pages={146--151},
  year={1997},
  organization={IEEE}
}

@article{thrun2002robotic,
  title={Robotic mapping: A survey},
  author={Thrun, Sebastian and others},
  journal={Exploring artificial intelligence in the new millennium},
  volume={1},
  number={1-35},
  pages={1},
  year={2002}
}

\end{document}